\documentclass[journal]{IEEEtran}
\usepackage{url}
\usepackage{multirow}
\usepackage{dirtree}
\usepackage{graphicx}
\usepackage{amsmath} 
\usepackage{xcolor}
\usepackage{hyperref}
\newcommand{\chA}[1]{{\color{black}{#1}}}

\begin{document}
\title{The LAIA Dataset: Labelled Attention for Intelligent Automobiles}

\author{{A. Contreras, D. Porres, R. Abad, P. Cano, A. Levy, G. Villalonga, A. M. López and A. Hernández-Sabaté}
\thanks{All the authors are with the Computer Vision Center. }
\thanks{A.M. López and A. Hernández-Sabaté are also with the Computer Science Department of Universitat Autònoma de Barcelona.}
}

\maketitle

\begin{abstract}
The development of autonomous vehicles (AVs) usually relies heavily on data-driven Artificial Intelligence (AI) models that require large volumes of sensor data with ground-truth annotations. While modular architectures are widely used, end-to-end driving paradigms offer a promising alternative by directly mapping sensor inputs to control actions. However, their adoption is limited due to challenges in interpretability and explainability. To address this, we present LAIA (Labelled Attention for Intelligent Automobiles), a novel synthetic dataset designed to enrich end-to-end driving research with human attention data. Collected using the CARLA simulator in closed-loop environments, LAIA comprises over 15 hours of driving from 44 participants across carefully crafted scenarios designed to evoke natural responses. Each sequence includes RGB images in six weather conditions, semantic and instance segmentation, depth, optical flow, CAN bus signals, and synchronized eye-tracking data. LAIA opens the doors to the use of human gaze driving for different applications such as training attention-aware end-to-end AI drivers, predicting driver behavior, or developing methods to detect anomalous driver-attention patterns, as well as to improve the explainability of the models. \chA{Within the scope of this work, we use LAIA to compare human attention with the perceptual attention emerging in our end-to-end driving models, thereby providing interpretability regarding their behavior.}
\end{abstract}


\IEEEpeerreviewmaketitle

\section{Introduction}

\chA{Autonomous vehicles could save hundreds of thousands of lives over the coming decades while reducing congestion, pollution, and travel time \cite{bauman2017why}, making autonomous driving one of the most active research areas in artificial intelligence, computer vision, and robotics. Among the paradigms proposed for vehicle automation, end-to-end driving has attracted growing attention for its direct mapping from sensory input to control actions, offering a scalable alternative to modular pipelines in which perception, prediction, planning, and control are treated as separate components. However, despite competitive driving performance, end-to-end models remain difficult to interpret \cite{chen2024end}, limiting their adoption relative to modular approaches whose explicit intermediate representations naturally support explainability. In safety-critical settings, predictive accuracy alone is insufficient: it is equally important to verify which scene elements a model attends to and whether they are the same elements that a human driver would consider relevant, such as pedestrians, traffic lights, or nearby vehicles. This challenge is rooted in how end-to-end models are trained — through behavior cloning from sensor-action pairs collected in human-driven vehicles — which captures what the driver did but not what the driver looked at, creating a gap between observable behavior and the underlying cognitive process. Human attention offers a promising way to narrow this gap: gaze data can serve both as an auxiliary supervisory signal during training, improving robustness and generalization, and as a post-hoc interpretability tool for evaluating whether model attention aligns with human visual decision-making.}

\begin{figure*}[h!t]
  \centering
  \includegraphics[width=0.9\textwidth]{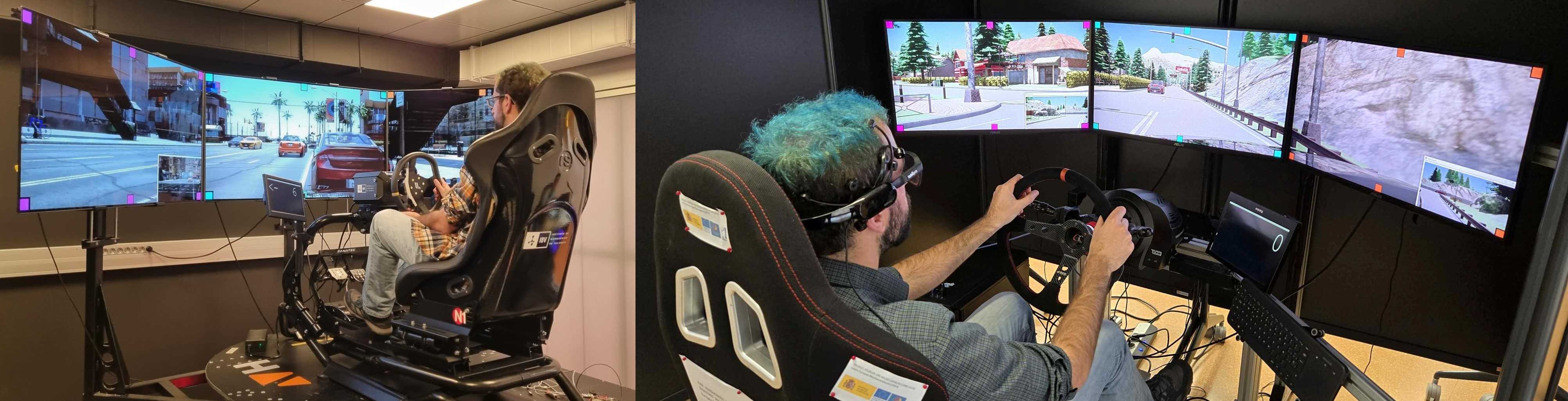}
  \caption{Platform Setup. {\it Left}: Dynamic driving platform. {\it Right}: Static driving platform. In both cases the simulation software is CARLA and the simulations used in human driving were based on the scenarios in Figure \ref{fig:events}.}
  \label{fig:simulator_setup}
\end{figure*}

We introduce {\bf LAIA} (Labelled Attention for Intelligent Automobiles), a public collection that couples eye gaze, control actions and full simulator annotations for large-scale autonomous-driving research, enabling reproducible large-scale studies of vision-guided driving. LAIA is designed to support the training, testing, and validation of end-to-end driving models while also enabling attention-based explainability studies. The dataset combines synchronized driving data with human attention information, allowing direct comparison between model attention and human gaze. By providing both the signals required for end-to-end learning and the information needed to analyze visual decision-making, the dataset aims to facilitate research toward autonomous driving systems that are not only accurate, but also more interpretable and cognitively grounded.

In particular, LAIA contains comprehensive driving scene data that reflect the inputs used by end-to-end AI driving models, including on-board sensor streams and vehicle control signals, all synchronized with human driver attention data captured via eye-tracking technology. The data were collected in closed-loop driving simulation environments, where 44 participants drove through a series of carefully designed routes featuring sudden and unexpected driving events. The simulation has been conducted using the CARLA autonomous driving simulator \cite{dosovitskiy2017carla}, complemented by a monitoring system that tracks human behavior while navigating within CARLA’s virtual towns. A total of 8 different CARLA routes with 18 different events carefully distributed to cause human reactions during their driving were implemented. RGB images in six environmental variations, along with ground-truth annotations for semantic segmentation, scene depth, panoptic instance segmentation, and synchronized human eye-tracking data are publicly released\footnote{https://cloningdcb.org/} \cite{DATA2570_2025}.

Unlike existing driving datasets, LAIA uniquely integrates dynamic human attention signals into the driving loop, providing a foundation for more cognitively grounded development of AI drivers. The dataset allows researchers to assess not only if an AI driver makes the correct \chA{driving} decisions, but \chA{also how similar its perception is with regard to human attention, an important step toward building interpretable and reliable autonomous systems}. Additionally, the attention maps extracted from human gaze data can serve as weak supervision or as an auxiliary task in multi-objective learning, supporting training strategies that improve both accuracy and robustness. Beyond learning, LAIA facilitates systematic benchmarking of attention-aware models and allows direct comparisons between human and model behavior under controlled, repeatable conditions.

The remainder of the paper is as follows. Section \ref{sc:Acquisition} describes all the elements necessary for creating the dataset, including the data recording protocol. Section \ref{sc:Preparation} is devoted to explaining the synchronization and mapping of the gaze data with the images generated by the simulator. Section \ref{sc:Format} describes the format and structure of the data while, Section \ref{sc:Examples} shows different examples of the use of the dataset. Finally, conclusions are summarized in Section \ref{sc:conclusion}.

\section{Data Acquisition Description}\label{sc:Acquisition}

Data were collected in two different centres, the Institute of Biomechanics of Valencia\footnote{www.ibv.org} (IBV) and the Computer Vision Center\footnote{www.cvc.uab.cat} (CVC), but the only differences among the sub-datasets lie in the physical simulator environments.

During the human driving sessions, eye-tracker data, CARLA images and additional data were captured to replay the episodes offline with the aim of generating additional data without preventing real-time performance during human driving sessions. Adapted autopilots accessing CARLA insider information were also used to automatically generate data associated with driving episodes. In all cases, erratic data were removed. 

In this Section, the main characteristics of the simulation environments, experimental scenarios (design, implementation, and experiment structure), and participants used to collect the database are described.

\subsection{Physical Simulation Environments}

The setup of the simulator environments consists of a system emulating a car, containing Hardware and Software Interfaces.

The simulator at IBV is mounted on a dynamic platform (Motion Systems PS-6TM-550) with a regulable seat that simulates the kinematic experience of a real vehicle (acceleration, braking, cornering, crashes and terrain irregularities) in response to manoeuvres executed by the human driver (Figure \ref{fig:simulator_setup}, left). The driving scene is displayed across three 55" 4K screens arranged at 46-degree angles between them.

The hardware simulated environment mounted at CVC is a static platform composed of a foldable aluminium frame, a Z-1 sports seat, the Thrustmaster T300 RS haptic simulation hardware including a GP-4 carbon offset steering wheel together with compatible turn signals and pedals (Figure \ref{fig:simulator_setup}, right). In this case, three articulated monitors of 28" and a resolution of 1920x1080, arranged at 46-degree angles between them, represent the front view of the car plus the two mirrors. In both simulators, a tablet displays the speed. 

CARLA simulator, version 0.9.14 was used to design and run the scenarios containing sudden events. Experiments were run on a PC (ProArt X670E-creator wifi, Ryzen-9 7900X, 2 x 32GB  PC5-5600U) with Windows 10 and NVIDIA GeForce 4090 Liquid plus NVIDIA GeForce 4090 INNO3D.

\begin{figure*}[h!t]
    \centering
   \begin{tabular}{|p{0.3\textwidth}p{0.3\textwidth}p{0.3\textwidth}|}
    \hline 
    \includegraphics[width=0.3\textwidth]{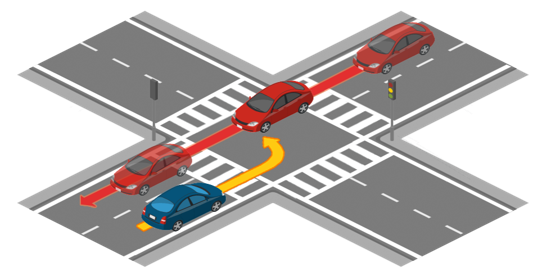} & \includegraphics[width=0.3\textwidth]{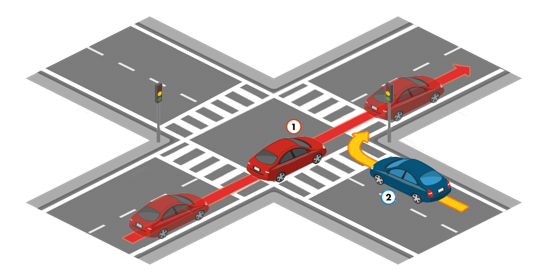} & \includegraphics[width=0.3\textwidth]{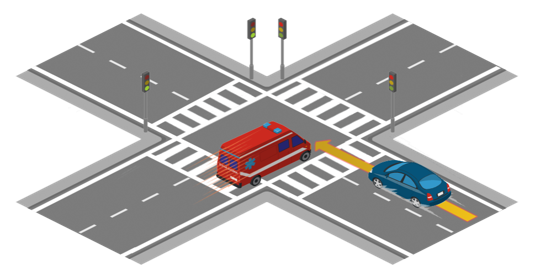} \\
    E2. Unprotected left turn at intersection & E3. Turn at intersection with crossing traffic & E5. Crossing traffic running a red light at an intersection \\
    
    \includegraphics[width=0.3\textwidth]{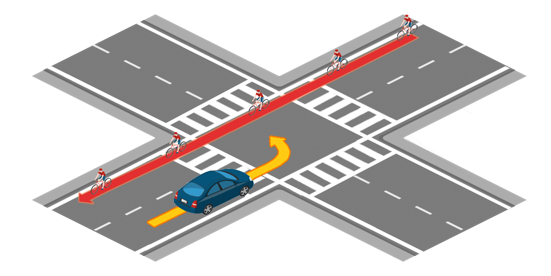} & \includegraphics[width=0.3\textwidth]{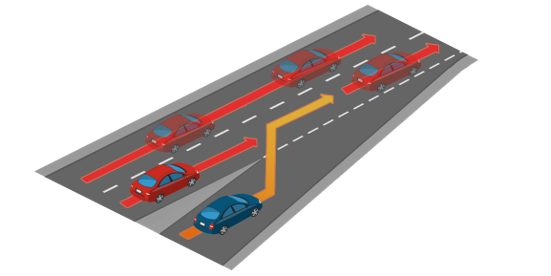} & \includegraphics[width=0.3\textwidth]{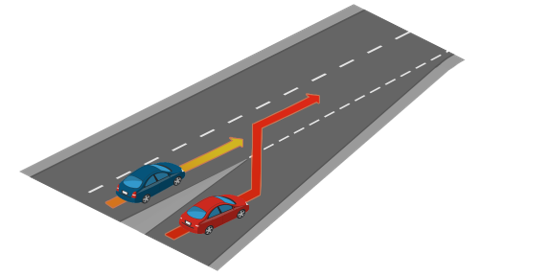} \\
    E6. Crossing with oncoming bicycles & E7. Highway merge on ramp & E8. Highway cut-in from on-ramp \\
    \includegraphics[width=0.3\textwidth]{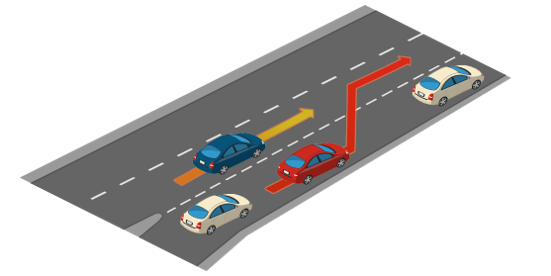} & \includegraphics[width=0.3\textwidth]{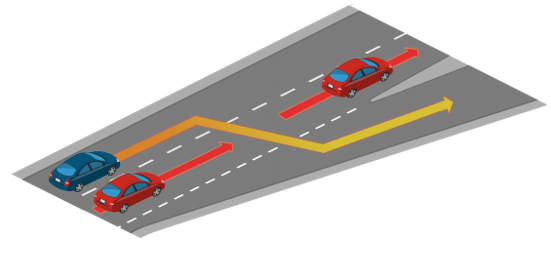} & \includegraphics[width=0.3\textwidth]{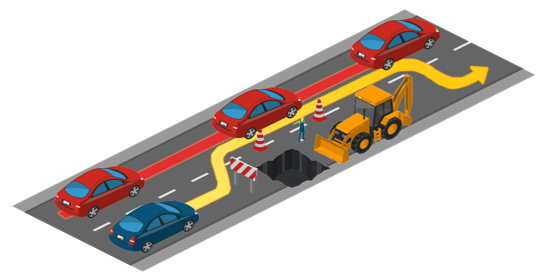} \\
    E9. Static cut-in & E10. Highway exit & E12. Obstacle in lane \\
    \includegraphics[width=0.3\textwidth]{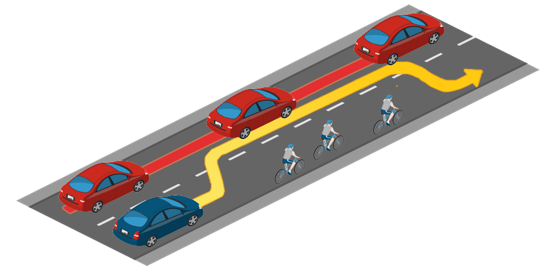} & \includegraphics[width=0.3\textwidth]{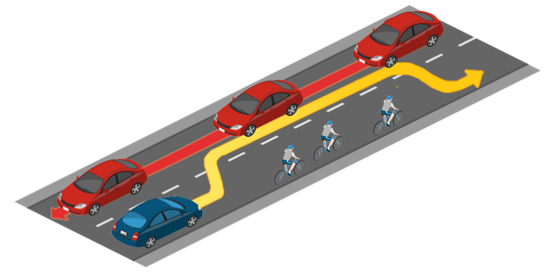} & \includegraphics[width=0.3\textwidth]{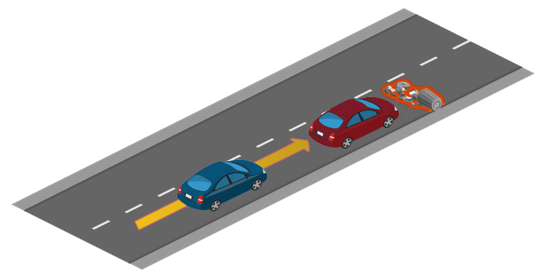} \\
    E14. Slow moving hazard at lane edge & E14a. Slow moving hazard at lane edge & E16. Longitudinal control after leading vehicle’s brake \\
    \includegraphics[width=0.3\textwidth]{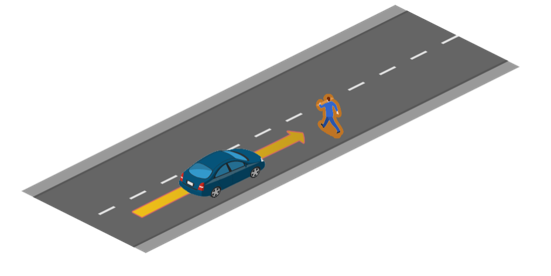} & \includegraphics[width=0.3\textwidth]{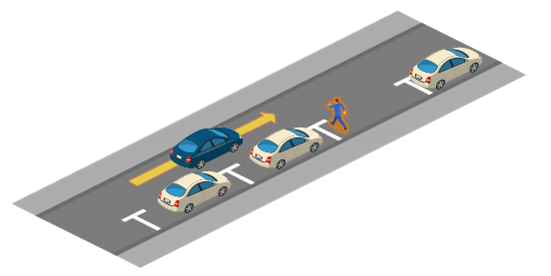} & \includegraphics[width=0.3\textwidth]{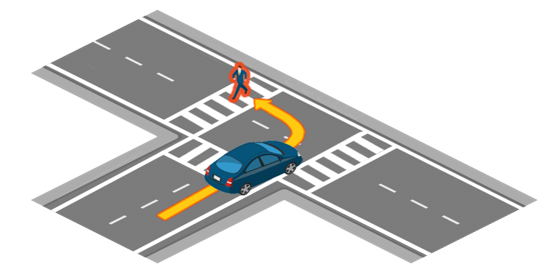} \\
    E17. Obstacle avoidance without prior action & E18. Pedestrian emerging from behind parked vehicle & E19. Obstacle avoidance with prior action - pedestrian or bicycle \\
    \includegraphics[width=0.3\textwidth]{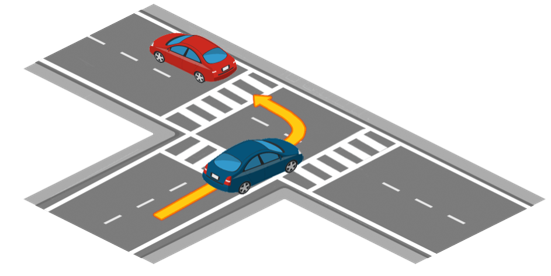} & \includegraphics[width=0.3\textwidth]{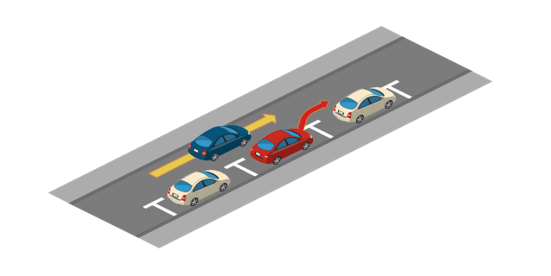}& \includegraphics[width=0.3\textwidth]{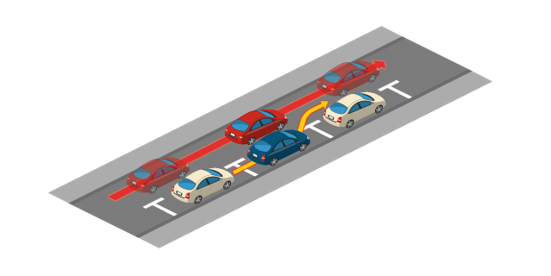}\\
    E19a. Obstacle avoidance with prior action - vehicle & E20. Parking cut-in & E21. Parking exit \\
    \hline
\end{tabular}
\caption{Implemented events.}
\label{fig:events}
\end{figure*}
\begin{table*}[h!t]
\centering
\caption{Event distribution across scenarios (see the event ID in fig. \ref{fig:events})}
\begin{tabular}{|l|cccc|cccc|ccc|ccccc|cc|}
\hline 
& \multicolumn{4}{c|}{\textbf{Traffic Negotiation}} & \multicolumn{4}{c|}{\textbf{Highway}} & \multicolumn{3}{c|}{\textbf{Obstacle Avoidance}} & \multicolumn{5}{c|}{\textbf{Braking and Lane Changing}} & \multicolumn{2}{c|}{\textbf{Parking}} \\ \hline 
& \textbf{E2} & \textbf{E3} & \textbf{E5} & \textbf{E6} & \textbf{E7} & \textbf{E8} & \textbf{E9} & \textbf{E10} & \textbf{E12} & \textbf{E14} & \textbf{E14a} & \textbf{E16} & \textbf{E17} & \textbf{E18} & \textbf{E19} & \textbf{E19a} & \textbf{E20} & \textbf{E21} \\ \hline
\textbf{Town01\_R01}      & x & x &   & x &   &   &   &   &   &   & x & x & x & x & x & x &   &   \\ \hline
\textbf{Town04\_R01}      & x & x & x &   &   &   &   &   &   &   &   & x &   &   & x &   &   &   \\ \hline
\textbf{Town04\_R02}      &   &   &   &   & x & x &   & x & x &   &   &   &   &   &   &   &   &   \\ \hline
\textbf{Town04\_R03}      &   &   & x & x &   &   &   &   &   &   &   &   & x &   &   &   &   &   \\ \hline
\textbf{Town06\_R01}      &   &   &   &   & x &   &   & x &   &   &   & x &   &   &   &   &   &   \\ \hline
\textbf{Town06\_R02}      &   &   &   &   &   &   & x &   & x & x &   &   &   &   &   &   &   &   \\ \hline
\textbf{Town06\_R03}      &   &   &   &   &   &   & x &   & x & x &   &   &   &   &   &   &   &   \\ \hline
\textbf{Town10\_R01}      &   &   &   &   &   &   &   &   &   &   &   &   &   &   &   &   & x  & x \\ \hline 
\end{tabular}
\label{tb:scenarios}
\end{table*}

\subsection{Simulated scenarios}

We define an {\bf episode} as the human driving experience from the time the subject sits on the platform until the end of driving in a predefined scenario.

We define a {\bf scenario}, named by TownXX\_RYY, as the combination of a town and a route comprising several events properly triggered as each subject drives along the pre-defined route. The instructions to follow the route were given by an automated voice (options: English/Spanish/Catalan). 

To design the scenarios, four different CARLA maps, namely towns, were selected and, for each map, between one and three routes were designed. Different events that may occur both in a city and on a highway were taken into account in a total of four towns with Single Lane (SL), Multilane (ML), or Highway (HW). In particular, the events implemented are the following and can be grouped in: 
\begin{itemize}
\item Traffic Negotiation:  Unprotected left turn at intersection, Turn at intersection with crossing traffic, Crossing traffic running a red light at an intersection and Crossing with oncoming bicycles.
\item Highway: Highway merge on ramp, Highway cut-in from on-ramp, Static cut-in, Highway exit.
\item Obstacle Avoidance: Obstacle in lane, Slow moving hazard at lane edge, Slow moving hazard at lane edge.
\item Braking and Lane Changing: Longitudinal control after leading vehicle’s brake, Obstacle avoidance without prior action, Pedestrian emerging from behind parked vehicle, Obstacle avoidance with prior action - pedestrian or bicycle, Obstacle avoidance with prior action - vehicle.
\item Parking: Parking cut-in, Parking exit.
\end{itemize}

Figure \ref{fig:events} illustrates all the events implemented while Table \ref{tb:scenarios} summarizes the distribution of the different events across the different scenarios, taking into account that SL is present in Towns 1 and 4, ML is present in towns 4 and 6 and HW can be found in towns 4, 6 and 10.

\begin{figure}[h!]
  \centering
  \includegraphics[width=0.9\linewidth]{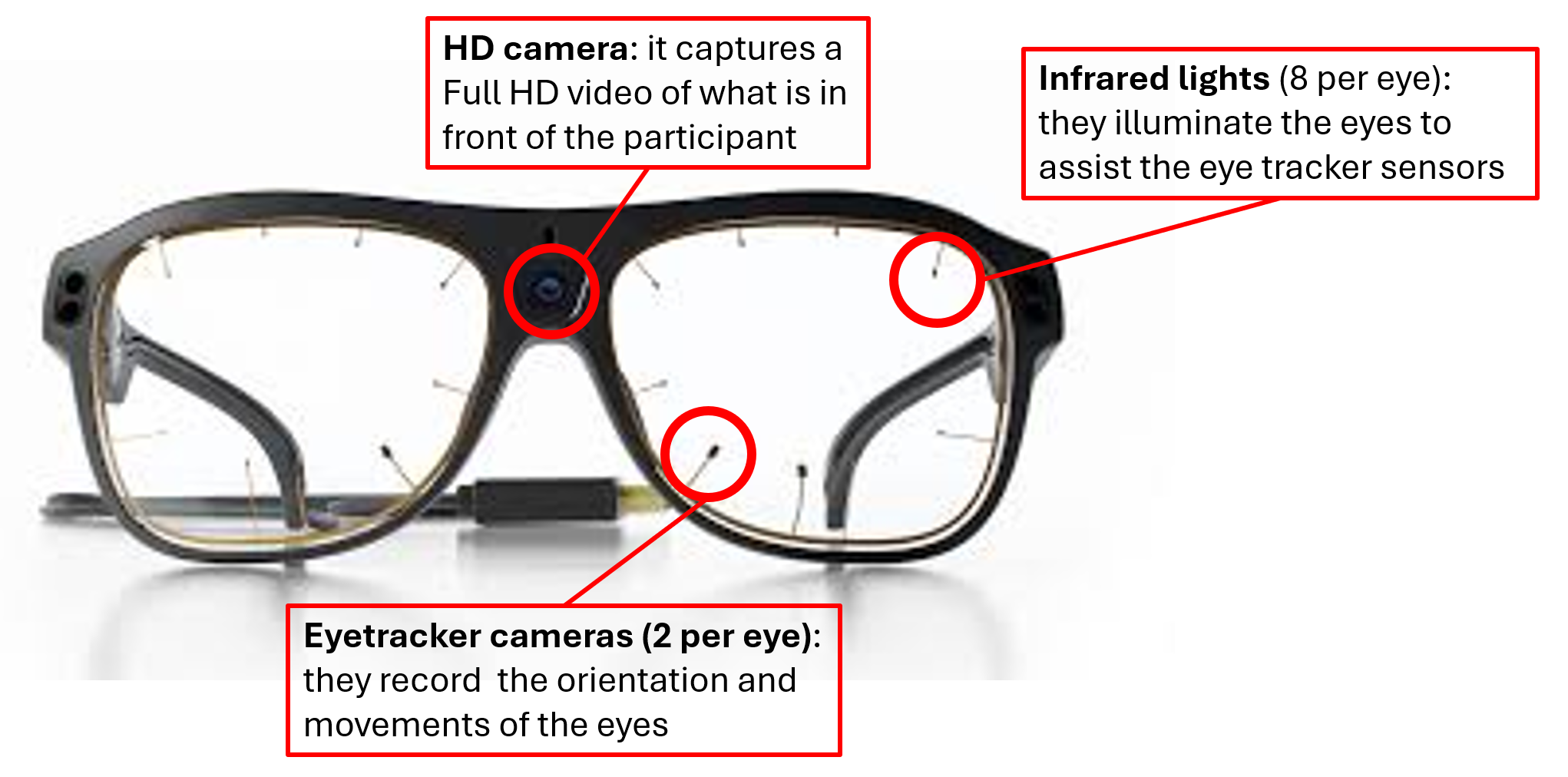}
  \caption{Eyetracker device.}
  \label{fig:eyetracker}
\end{figure}

\begin{figure*}[h!]
    \centering
   \includegraphics[width=0.9\textwidth]{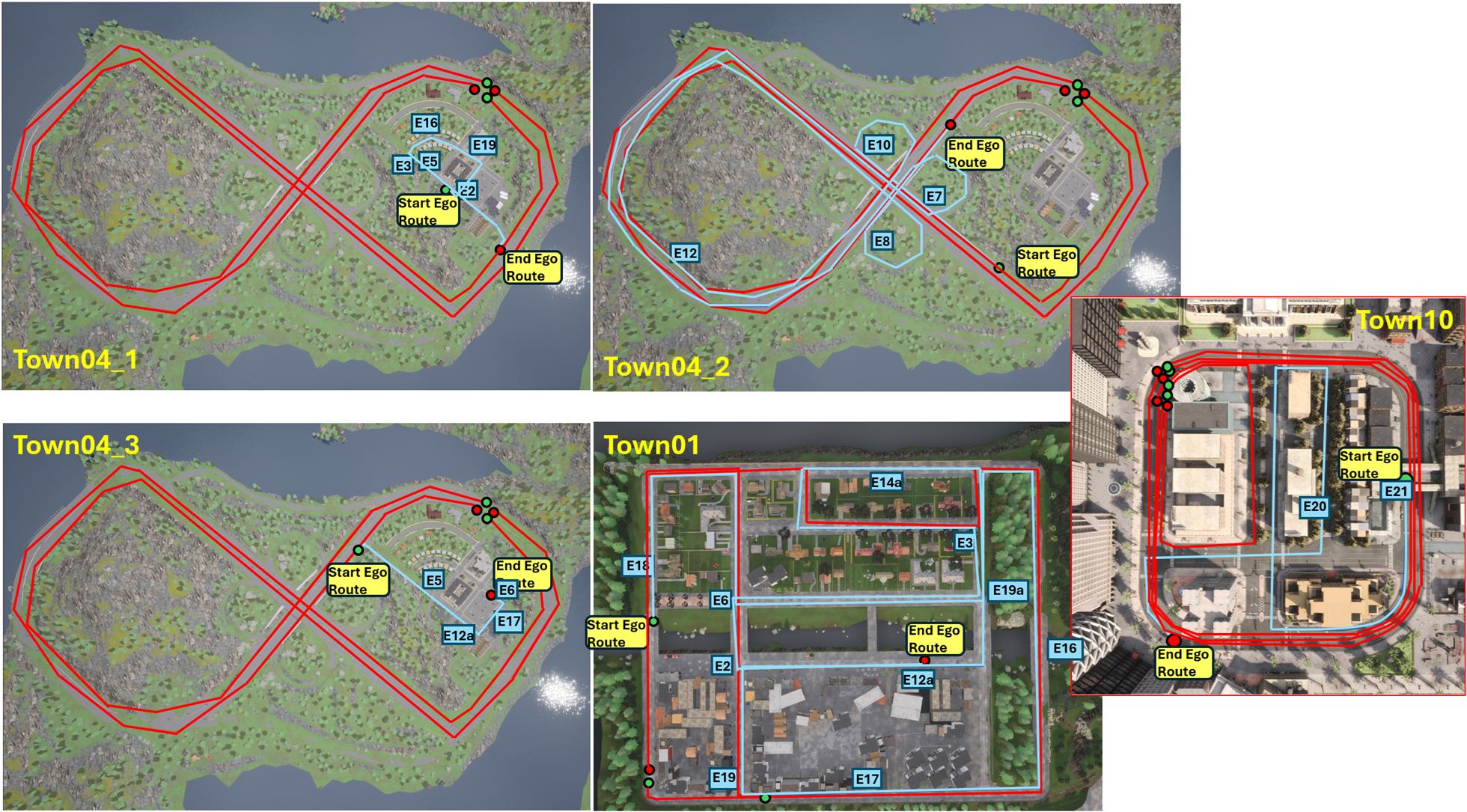} \\   
\caption{Scenarios for CARLA's towns 01, 04 and 10. Blue lines show the route the human drivers have to follow, from the starting point in dark green to the ending red one. Red lines show the automatic vehicles trajectories, from the starting bright green points to the smallest red ones. Events are marked with a blue box.}
\label{fig:maps1}
\end{figure*}
\begin{figure*}[h!]
    \centering
   \includegraphics[width=0.9\textwidth]{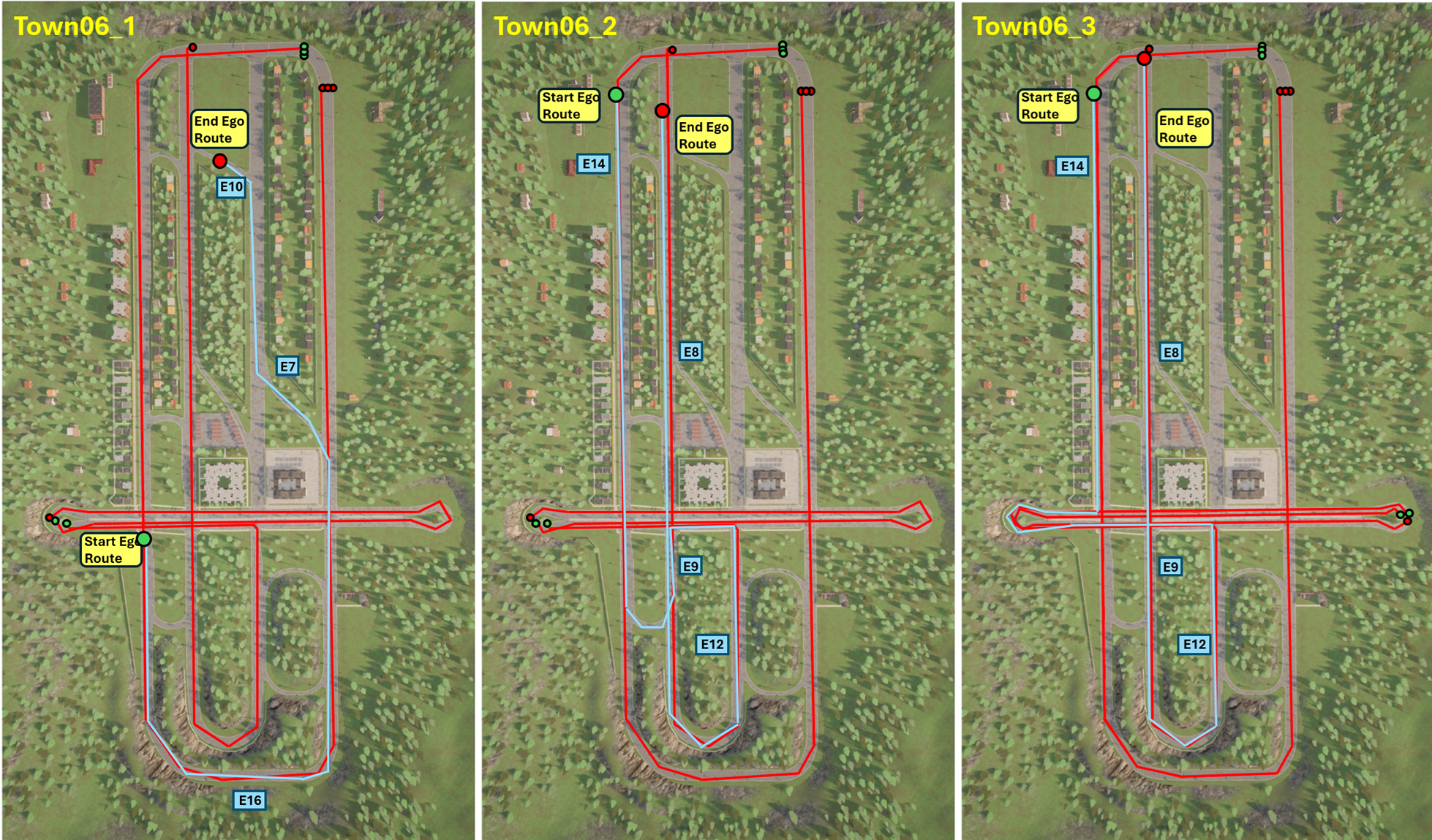} \\   
\caption{Scenarios for town maps 06. Blue lines show the route the human drivers have to follow, from the starting point in dark green to the ending red one. Red lines show the automatic vehicles trajectories, from the starting bright green points to the smallest red ones. Events are marked with a blue box.}
\label{fig:maps2}
\end{figure*}
For a better illustration Figures \ref{fig:maps1} and \ref{fig:maps2} show the different CARLA maps with the routes the human drivers have to follow, the simulated vehicles trajectories and the points where the events are implemented.

\subsection{Eyetracker}
To capture driver attention a wearable eye-tracker, as the one shown in Figure \ref{fig:eyetracker}, was used (Tobii glasses 3 \cite{tobii_pro_glasses_3}). These glasses are equipped with four eye cameras and a wide-angle scene camera, allowing for precise tracking of eye movements and visual attention. The glasses are lightweight and unobtrusive, ensuring natural behavior from participants during studies. Gaze data was recorded at 100Hz and the scene camera recorded at 25 fps with a resolution of 1920 x 1080 pixels.

\subsection{Participants}

A total of 44 participants (23 women and 21 men) were asked to drive during 20-30 minutes across 3-4 scenarios. To ensure a baseline level of driving competence and familiarity with common traffic situations, participants were healthy people without any condition that might have caused an imbalance in the data recorded:
\begin{itemize}
    \item At least two years of driving license.
    \item Normal or corrected to normal vision (without glasses).
    \item Very little tendency to suffer motion sickness.
    \item Non-professional drivers.
    \item Driving at least once a week.
\end{itemize}
\chA{Professional drivers were explicitly excluded to avoid over-representation of highly automated or task-optimized driving behaviors that may not reflect typical human attention patterns.} The average age of participants was $36.59 \pm 9.31$ and the average of years of license was $16.23 \pm 8.76$. 
The characteristics of all the participants regarding age, years of license and number of scenarios run are detailed in table \ref{tb:participants}:

\begin{table}[h!]
\centering
\caption{Characteristics of participants}
\begin{tabular}{ll|c|c|c|c|}
\cline{3-6}
                                           &       & \# Drivers & Age            & License         & Scenarios          \\ \hline
\multicolumn{1}{|l|}{\multirow{2}{*}{IBV}} & Men   & 11             & 40 $\pm$ 5.99  & 20 $\pm$ 6.82   & \multirow{2}{*}{3} \\ \cline{2-5}
\multicolumn{1}{|l|}{}                     & Women & 13             & 39 $\pm$ 9.02  & 19 $\pm$ 9.21   &                    \\ \hline
\multicolumn{1}{|l|}{\multirow{2}{*}{CVC}} & Men   & 10             & 31 $\pm$ 12.2  & 12.4 $\pm$ 10.7 & \multirow{2}{*}{4} \\ \cline{2-5}
\multicolumn{1}{|l|}{}                     & Women & 10             & 32.4 $\pm$ 9.5 & 11.9 $\pm$ 8.6  &                    \\ \hline
\end{tabular}
\label{tb:participants}
\end{table}

The details of recordings per scenario are detailed in table \ref{tb:recordings}. Notice that the number of frames for each modality is the same because they are synchronized.

\begin{table}[h!]
\centering
\caption{Recording details}
\begin{tabular}{l|c|c|c|}
\cline{2-4}
   \multicolumn{1}{l|}{}         & \# drivers & \# frames & Total time (minutes) \\
\hline
\multicolumn{1}{|l|}{Town01\_R01} & 38                & 667211                 & 445.19                    \\
\hline
\multicolumn{1}{|l|}{Town04\_R01} & 10                & 58397                  & 38.96                   \\
\hline
\multicolumn{1}{|l|}{Town04\_R02} & 14                & 132884                 & 88.65                   \\
\hline
\multicolumn{1}{|l|}{Town04\_R03} & 13                & 88725                  & 59.19                   \\
\hline
\multicolumn{1}{|l|}{Town06\_R01} & 12                & 37119                  & 24.76                     \\
\hline
\multicolumn{1}{|l|}{Town06\_R02} & 11                & 89349                  & 59.61                   \\
\hline
\multicolumn{1}{|l|}{Town06\_R03} & 9                 & 66601                  & 44.43                   \\
\hline
\multicolumn{1}{|l|}{Town10\_R01} & 27                & 209468                 & 139.75                   \\
\hline

\end{tabular}
\label{tb:recordings}
\end{table}

\subsection{Data Recording Protocol}

The main steps of the recording protocol are the following:

\begin{enumerate}
\item The participant receives the information sheet about the project under which LAIA is developed and the informed consent to be signed.

\item \chA{As soon as the participant agrees, (s)he sits in the driving simulator and puts on the eye-tracker glasses.}

\item \chA{The participant is asked to drive in the simulated environment, relying on the driving simulator and following the audio instructions for approximately half an hour.} These instructions are used to guide the driver along the specified route, in the same way as commercial navigation assistants do.
\item During human driving, the following data is collected:
\begin{enumerate}
    \item Driver's gaze captured by the eye-tracking glasses.
    \item The images projected by CARLA to provide the driving experience.
    \item Log of the driving simulation to allow offline playback of the driving episodes. In particular, CARLA data is captured using adapted autopilots that access all the CARLA privileged information needed for expert-level driving. That is, the autopilots know where the road is, lane lines, cars, pedestrians, traffic lights, etc. without the need to interpret images or any other sensory information.
\end{enumerate}
\end{enumerate}

The methodology of the experimentation procedure was approved by the "Comitè d'Ètica en la Recerca (CERec) de la Universitat Autònoma de Barcelona" (research ethics committee of the university, reference number: CEEAH 6936). For all the experiments, a written informed consent was obtained from each participant. The consent form explains the goal of the experiment and describes what kind of data are collected and the terms of privacy in the use of personal data. Additionally, it emphasizes that the data released to the general public does not contain information that can directly identify the subject and that any data and research results already shared with other investigators or the general public cannot be destroyed, withdrawn, or recalled. Each consent was hand-signed by each subject on the day of the first experiment. 

The eye-tracking glasses were manually calibrated for each driver prior to the start of every scenario. Both human driving data and data captured by autopilots were reviewed to eliminate episodes of erratic driving.

\section{Data Preparation and Validation}\label{sc:Preparation}
\begin{figure}[h!t]
  \centering
  \includegraphics[width=0.9\linewidth]{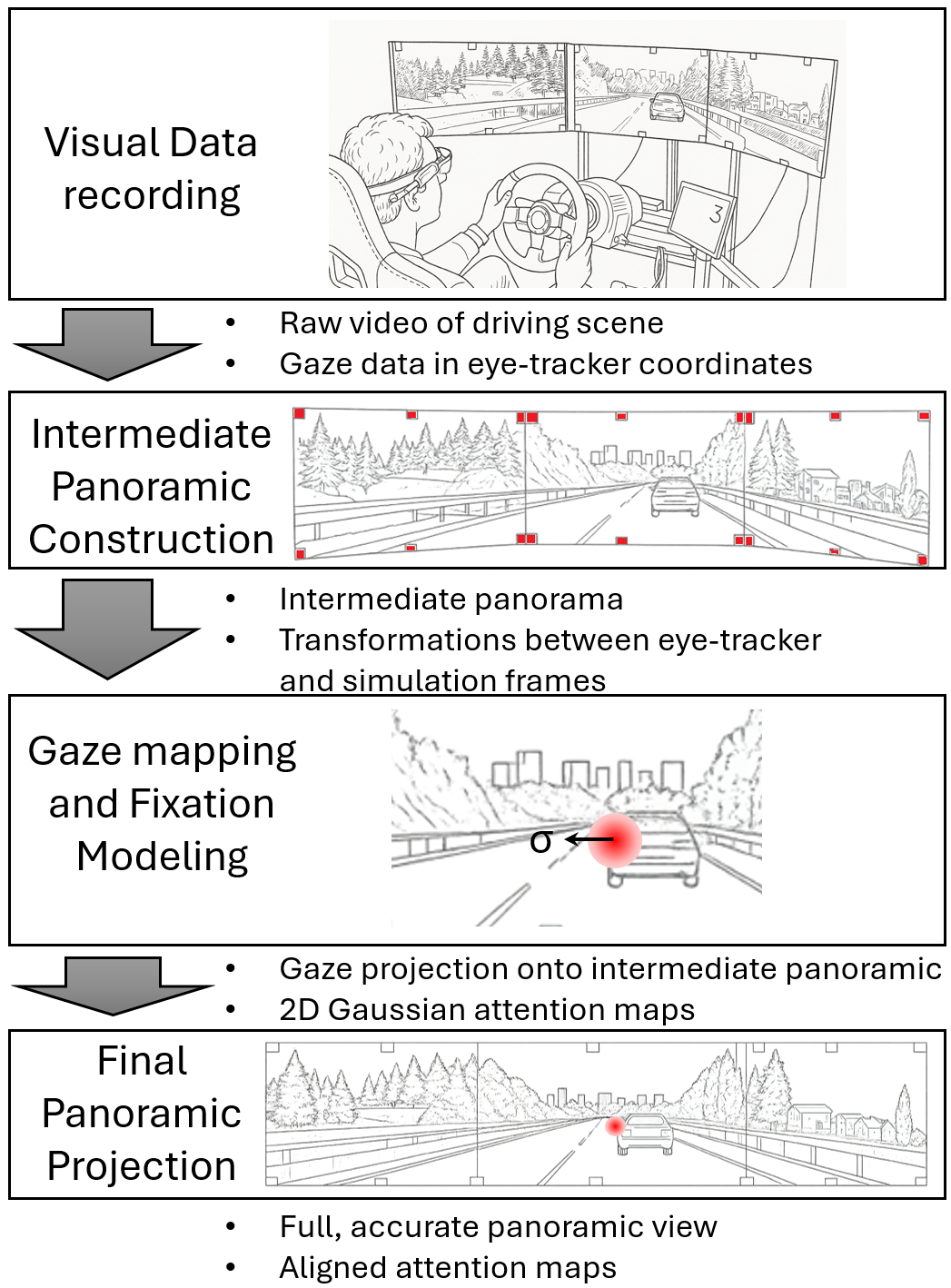}\\
  \caption{Attention mapping pipeline.}
  \label{fig:attention_pipeline}
\end{figure}

To synchronize the driver attention data with the visual driving context, the eye-tracking videos captured during driving sessions are mapped to high-fidelity panoramic imagery generated using the CARLA simulator’s playback functionality. It is worth noting that this process revealed a bug\footnote{https://github.com/carla-simulator/carla/pull/8960} in CARLA’s playback system, where, in some cases, spawning new actors caused ID conflicts, leading to incorrect aliasing of IDs during playback. \chA{We fixed the bug before producing the playbacks contained in LAIA. As figure \ref{fig:attention_pipeline} shows,} the attention mapping pipeline proceeds through the following stages: 1) Visual Data Recording, 2) Intermediate Panoramic Construction, 3) Gaze Mapping and Fixation Modeling, and 4) Final Panoramic Projection.

\subsection{Visual Data Recording}
\begin{figure}[h!t]
  \centering
  \includegraphics[width=0.9\linewidth]{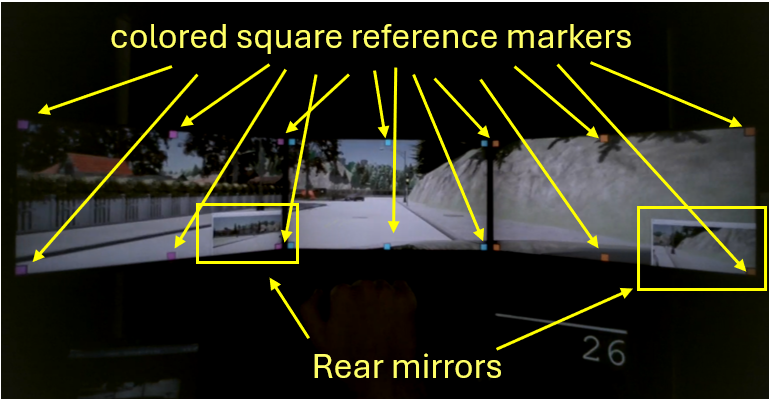}\\
  \caption{Original view of the eye tracker worn by the driver.}
  \label{fig:OriginalView}
\end{figure}

The raw visual data is recorded using eye-tracking glasses worn by the driver. As Figure \ref{fig:OriginalView} shows, the eyetracker’s scene camera records all simulator displays visible to the participant, including the rearview mirrors, which are positioned to replicate their real-world positions within the cockpit. 

Rendered square reference markers, with a unique color for each screen, are embedded into the borders of the simulator’s physical displays. These markers serve as fixed reference points to geometrically align the eyetracker video frames with the composite CARLA-rendered panorama. \chA{Their small size and simple shape were intentionally selected to reduce visual saliency and minimize the likelihood of interfering with the driver’s natural attention.} Narrow vertical strips at the junctions between displays are minimally visible in the eyetracker video and are considered negligible for alignment purposes.
\begin{figure*}[h!t]
  \centering
  \includegraphics[width=0.9\linewidth]{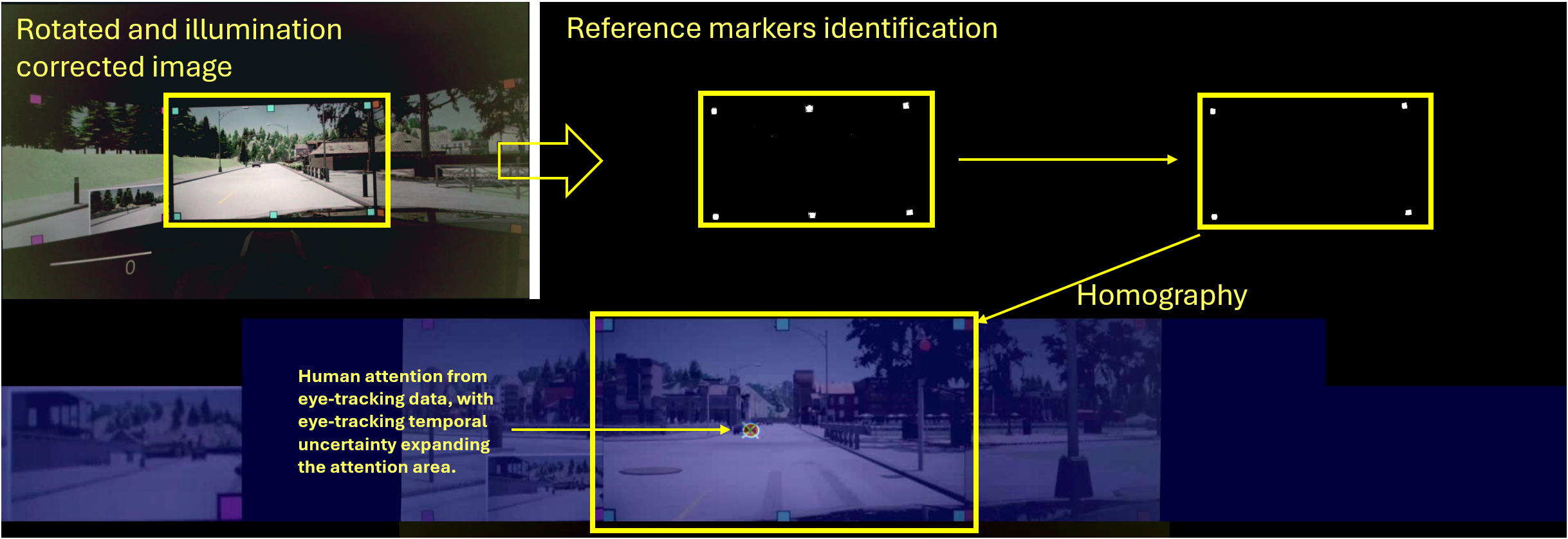}\\
  \caption{Intermediate Panoramic Construction.}
  \label{fig:PanoramicConstruction}
\end{figure*}

\subsection{Intermediate Panoramic Construction}
Using projective geometry and the aforementioned markers, an intermediate panoramic image is constructed offline to establish a common geometric frame between the eyetracker footage and the simulator imagery. First, each raw eyetracker image is rotation-corrected using head-tilt information from the glasses’ Inertial Measurement Unit (IMU), aligning the camera view with the horizontal axis of the display setup. Next, color normalization is applied to compensate for illumination differences between the eyetracker video and the CARLA rendering. The colored reference markers are then detected with a color-based filtering procedure, and their configuration within the screen layout of simulator display are identified by a geometric analysis. Using these correspondences, a projective mapping (homography) is estimated to warp gaze coordinates from each eyetracker view into an intermediate panoramic coordinate system extracted from the CARLA playback image. 

Figure \ref{fig:PanoramicConstruction} illustrates the mapping of an eyetracker frame into the intermediate panoramic coordinate system. For consistency with this panoramic playback image, the rearview mirrors are ultimately represented as the leftmost and rightmost sections of the resulting representation. Notice that this image spatially aligns the eyetracker’s visual field with the simulated environment but it contains only regions observed by the eyetracker and therefore excludes areas outside the eyetracker’s field of view. It is used solely as a staging space for robust, frame-accurate gaze projection. 


\subsection{Gaze Mapping and Fixation Modeling}
To project driver attention, gaze data are mapped from the eyetracker frame onto this intermediate panorama. \chA{Following the Tobii I-VT Attention Filter, which Tobii explicitly recommends for Tobii Pro Glasses 3 recordings in mobile environments,} a fixation is defined as a set of gaze points where the maximum angular velocity does not exceed 100 degrees per second and the duration ranges between 0.1 and 0.5 seconds. \chA{This higher threshold, relative to the 30°/s used in stationary lab studies, is necessary because head movements during driving inflate point-to-point velocities, causing lower thresholds to misclassify fixations as saccades \cite{hossain2016eye}. As a known trade-off, this threshold includes smooth pursuits and a small fraction (10–15\%) of short saccades, which slightly overestimates attention dwell time \cite{andersson2017one}.} Fixations lasting longer than 0.5 seconds are divided into shorter fixations.

Thus, attention is modeled as a two-dimensional Gaussian distribution centered at the gaze point, with the standard deviation $\sigma$ accounting for multiple sources of uncertainty:

\begin{itemize}
    \item The intrinsic fixation error of the eye-tracking device (typically $<5$ pixels in the eyetracker image).
    \item Downsampling from the $100$~Hz gaze stream to the $25$~fps scene camera.
    \item Temporal misalignment between eyetracker and simulator video playback.
    \item Reduction of human attention to a single gaze coordinate per frame.
\end{itemize}

Some regions remain without information in this intermediate image, as they fall outside the eyetracker's field of view. These areas are recovered in the next stage.

\subsection{Final Panoramic Projection}
The complete and accurate panoramic image is reconstructed offline using CARLA’s playback system, which renders all camera views (forward, lateral, rear, and mirrors) and can provide additional channels such as semantic labels, depth, optical flow, and CAN bus states. Because playback is driven by ground-truth CARLA simulator logs rather than the participant’s head pose, it produces a consistent, full-coverage panorama that includes regions not visible in the eyetracker footage.

\begin{figure*}[h!t]
  \centering
  \includegraphics[width=0.9\linewidth]{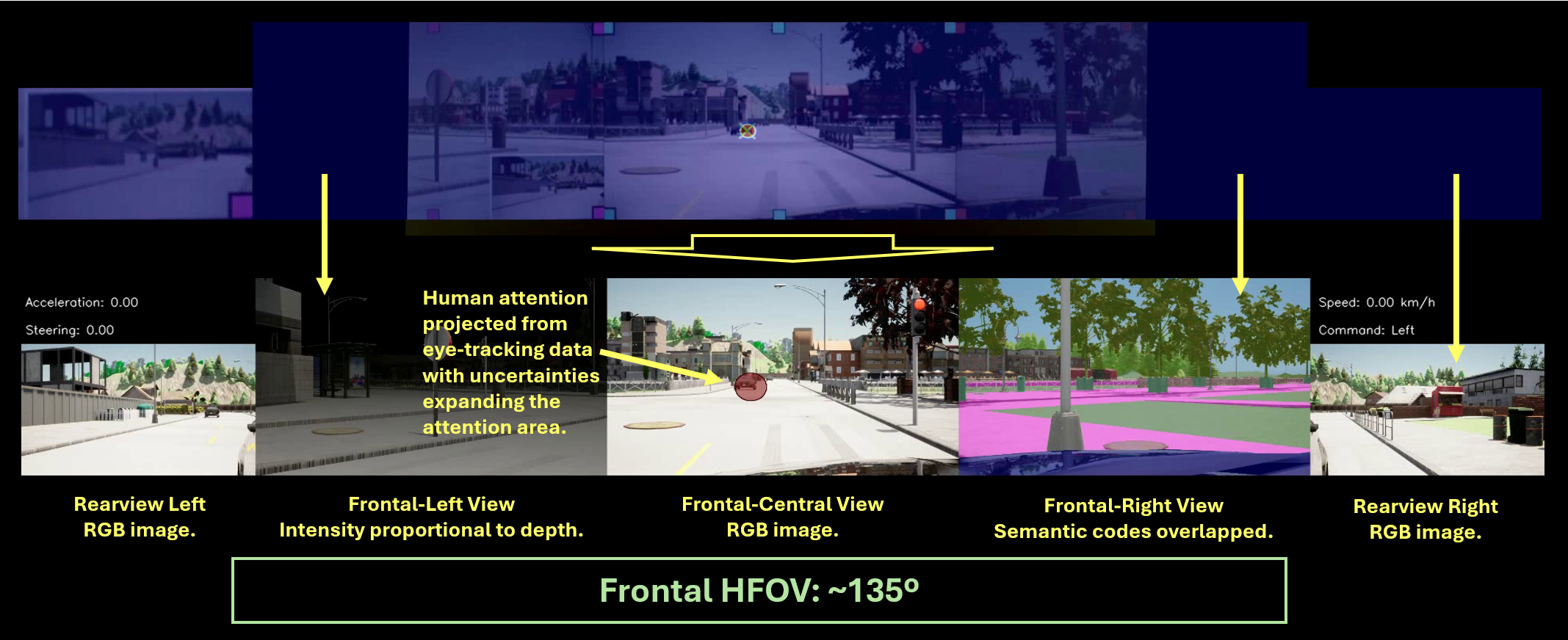}
  \caption{Final Panoramic Projection. On the top, the intermediate panoramic image using projective geometry. On the bottom the final panoramic image generated using CARLA's playback feature with an additional channel of semantic segmentation. Human attention is projected and expanded with uncertainty modeled by a Gaussian distribution.}
  \label{fig:FinalView}
\end{figure*}

The attention maps computed in the intermediate space are then projected into this final panoramic coordinate system, ensuring that gaze and attention are precisely aligned with the full simulator scene. The Gaussian attention fields retain their modeled uncertainty, thereby propagating spatial and temporal confidence through the entire pipeline. This rigorous multi-step process ensures that human attention is accurately reflected in the original simulator-based visual scenes. The final result is suitable for training, evaluation, and interpretability analyses of attention-aware driving models.

Figure \ref{fig:FinalView} illustrates the final process of the panoramic image reconstruction together with the correct projection of gaze data. Notice that missing information in the eyetracker video data is recovered in the final panoramic view. \chA{This figure also shows samples of the depth channel and the semantic segmentation channel in the Frontal-Left and the Frontal-Right views, respectively.}

\subsection{Qualitative Validation}
\begin{figure}[h!t]
  \centering
  \begin{tabular}{c}
  \includegraphics[width=0.9\linewidth]{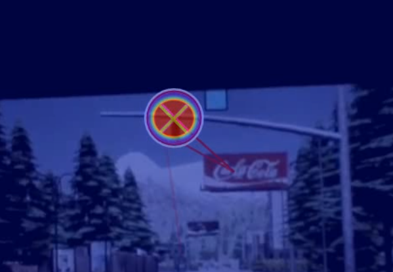}\\
  \includegraphics[width=0.9\linewidth]{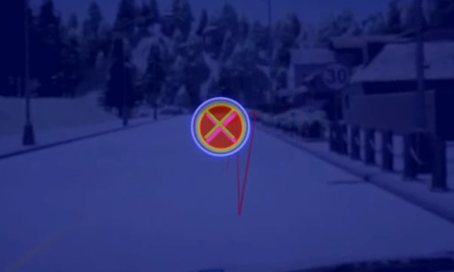}\\
  \end{tabular}
  \caption{Visualization of gaze uncertainty modeling and comparison to the original Gaussian provided by the eye-tracking device. \chA{The broader heatmap-like distribution corresponds to the native distribution provided by the eye-tracker device, while the red-outlined circles represent the attention distribution modeled by our pipeline.}}
  \label{fig:EyetrackingMappingDetails}
\end{figure}

To qualitatively support the correctness of our Gaussian modeling of gaze attention, Figure~\ref{fig:EyetrackingMappingDetails} compares two Gaussian-based representations of the driver gaze: \chA{
the native distribution provided by the eye-tracker (colored heatmap), together with the scanpath (red lines), and the distribution computed by our pipeline (red-outlined circle). The inner crosses mark the instantaneous gaze point recorded by the device, and the centers of both Gaussians are nearly perfectly concentric.}

\begin{figure*}[h!t]
  \centering
  \includegraphics[width=\linewidth]{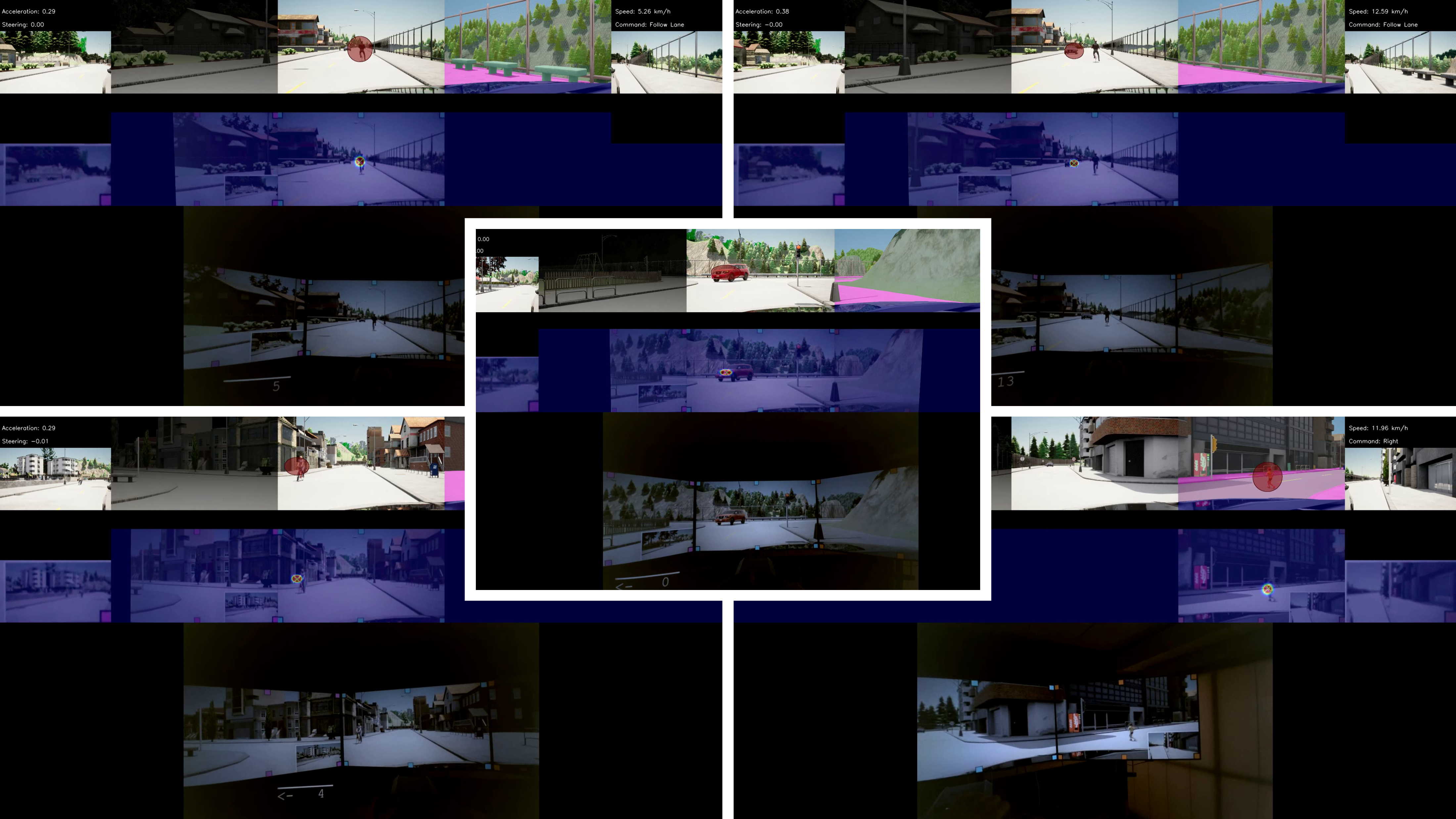}
  \caption{Qualitative validation of the eye-tracking mapping onto the original simulator-grounded visual scenes. Each grouped image contains the original image captured by the eye-tracking on the bottom, the intermediate panoramic image with a heat map visualizing the projected human attention, together with the instant gaze produced by the eye-tracker device, and the final panoramic image with the gaze data accurately projected.}
  \label{fig:EyetrackingMappingValidation}
\end{figure*}

Figure \ref{fig:EyetrackingMappingValidation} shows different examples of qualitative validation of the whole mapping. The widths of our Gaussians vary depending on the duration of the fixation.

\section{Data Format and Structure}\label{sc:Format}
The dataset described has been made publicly available in a federated and multidisciplinary data repository \cite{DATA2570_2025} and in  \href{https://cloningdcb.org/#loaded}{https://cloningdcb.org/}. No registration is required and users can select and download specific data through a customizable download configurator. This interface allows users to tailor their dataset selection based on their needs, ensuring flexible and efficient access to the data.

\begin{figure*}[h!t]
  \centering
  \begin{tabular}{cc}
  \includegraphics[width=0.5\linewidth]{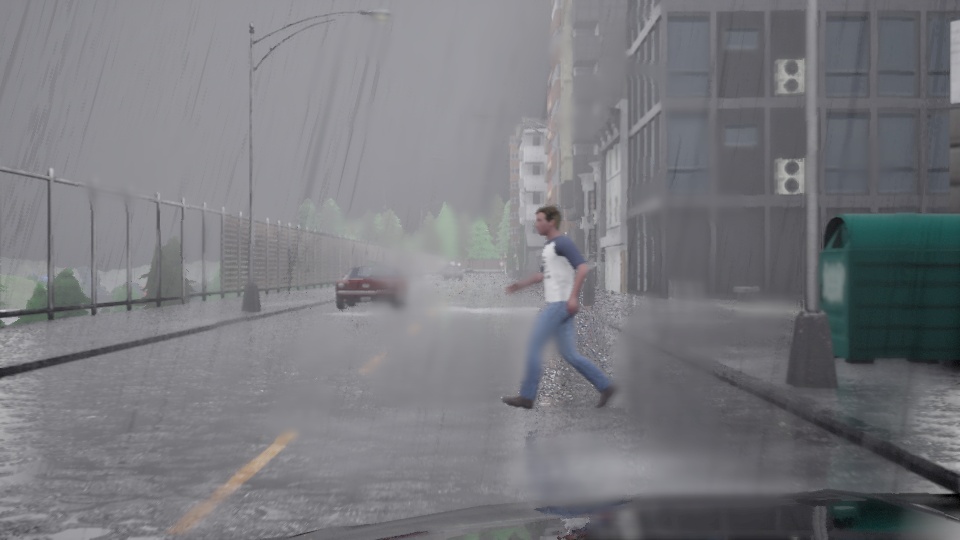} & \includegraphics[width=0.5\linewidth]{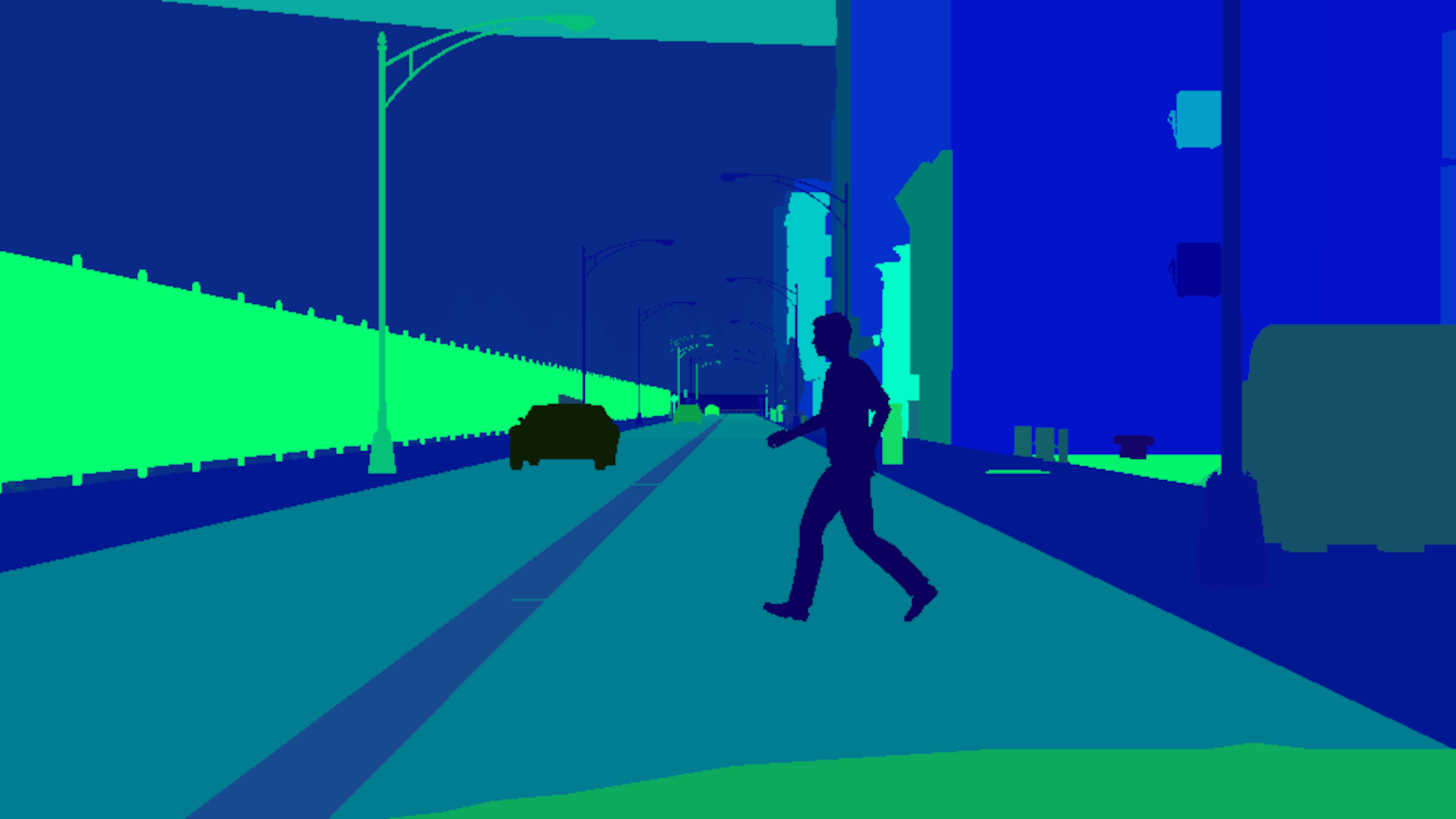}\\
  (a) & (b)\\
  \includegraphics[width=0.5\linewidth]{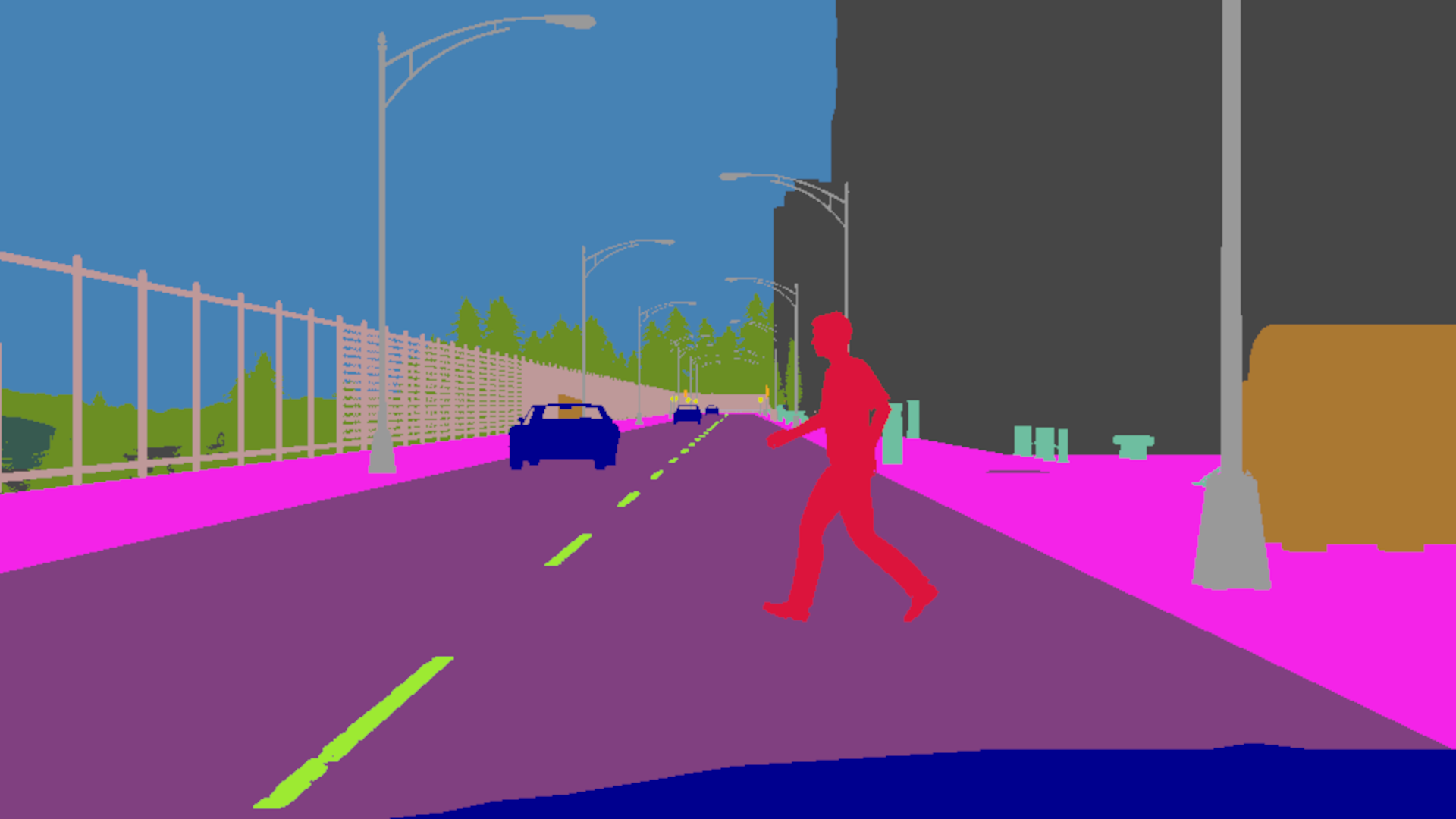} & 
  \includegraphics[width=0.5\linewidth]{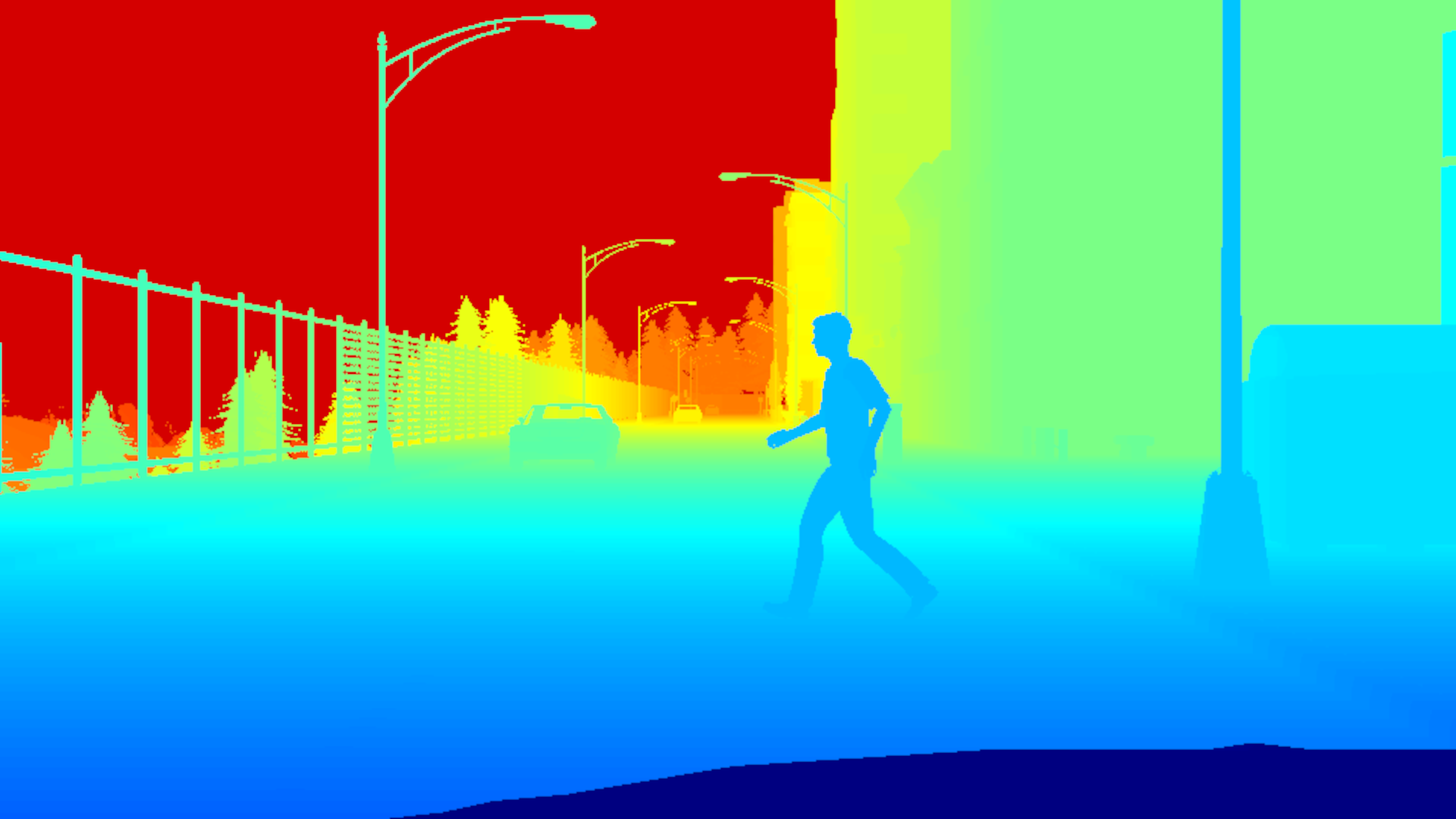} \\
  (c) & (d)\\
  \includegraphics[width=0.5\linewidth]{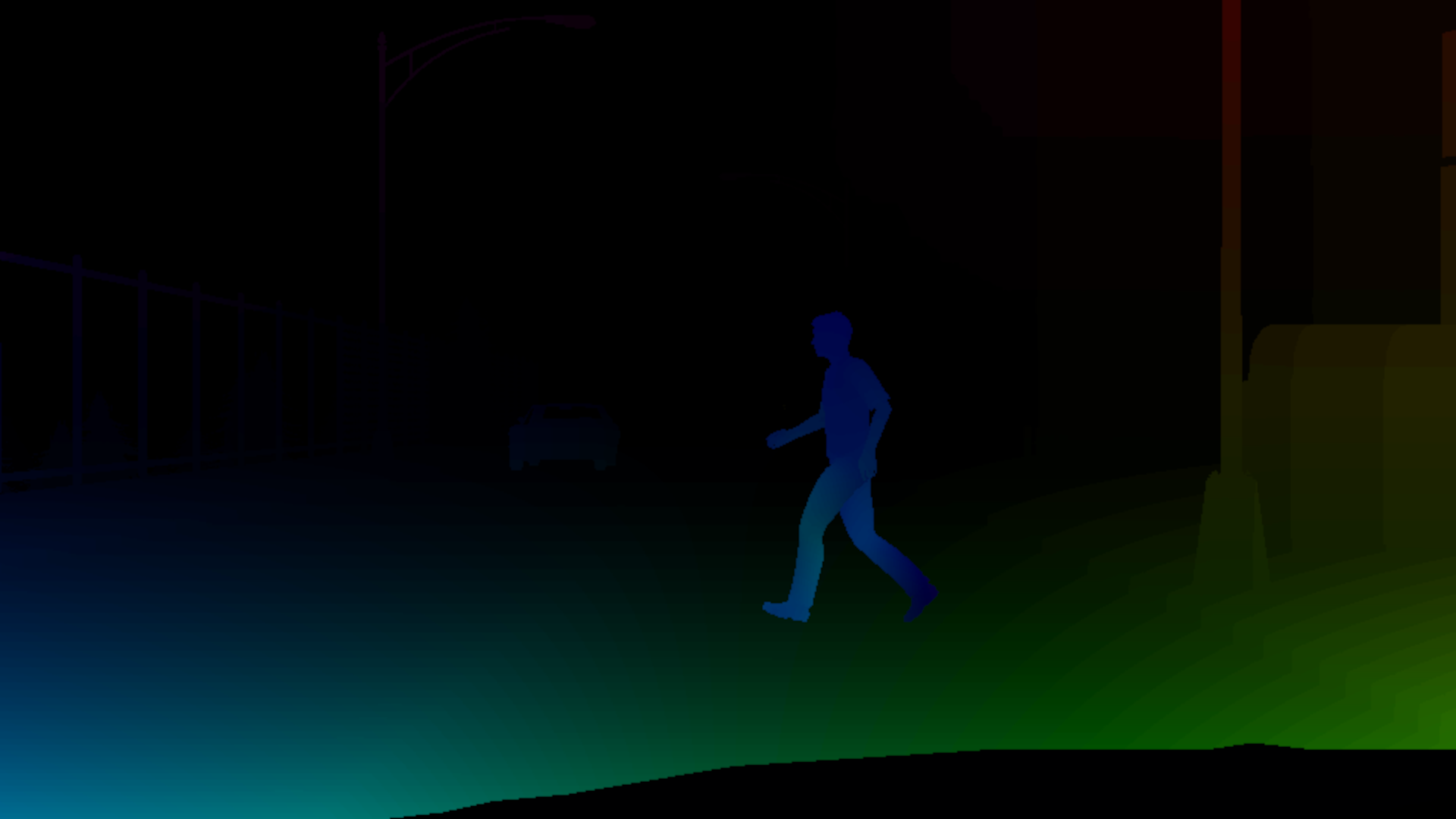} & 
  \includegraphics[width=0.5\linewidth]{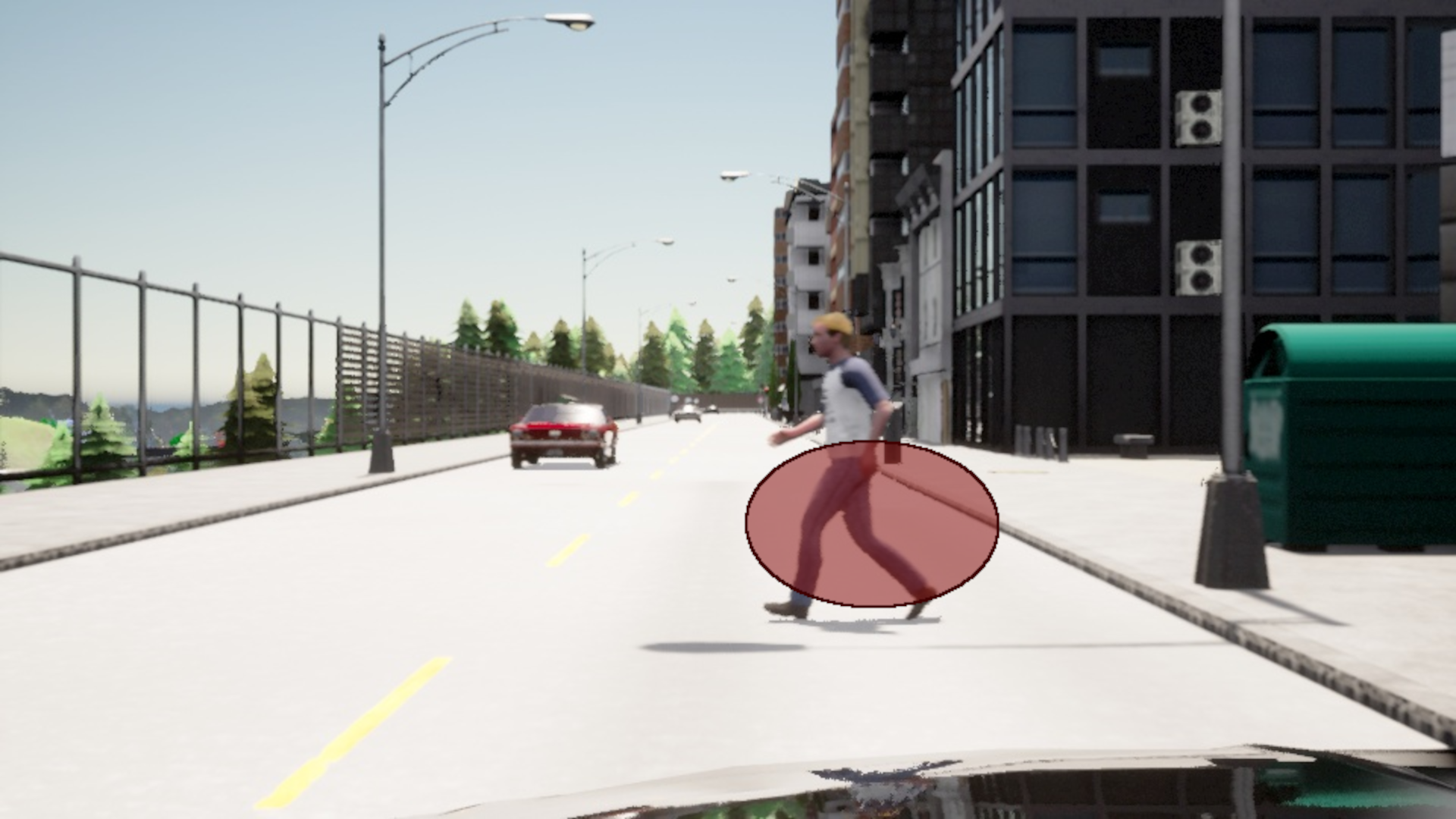} \\
  (e) & (f)\\
  
  \end{tabular}
  \caption{Illustration of available data. a) RGB , b) Panoptic Instance Segmentation, c) Panoptic Semantic Segmentation, d) Depth, e) Optical flow, and f) Gaze overlaid on an RGB image.}
  \label{fig:GT}
\end{figure*}

\chA{Below, we list the publicly available data, while Figure~\ref{fig:GT} presents representative examples of each data modality included in the dataset.}

For each driver and each scenario the data available are as follows: 
\begin{enumerate}
    \item \chA{Playback} RGB images, obtained from central, right and left cameras, as well as left and right mirrors \chA{(Fig~\ref{fig:GT}(a))}. These images are generated in 6 different weather conditions: clear noon, clear sunset, Hard rain noon, wet noon, cloudy noon, mid rain sunset.
    \item \chA{Panoptic Segmentation: instance (Fig~\ref{fig:GT}(b)) and semantic segmentation (Fig~\ref{fig:GT}(c)).}
    \item Depth (pixel distance obtained per camera using a CARLA depth camera) (Fig~\ref{fig:GT}(d)).
    \item Optical flow data (Fig~\ref{fig:GT}(e)).
    \item CAN bus information at every timestamp, at a frequency of 25 Hz. It contains a .json file with the information of speed, acceleration, brake, throttle, steer, ego position, gyroscope, accelerometer, hand\_brake, rear\_gear, blinker, and direction. All of these signals are synchronized with all the data.
    \item Attention map generated from eye-tracker mapped on the aligned RGB images (Fig~\ref{fig:GT}(f)).
\end{enumerate}
\newpage
Internally, data are structured as follows:
\dirtree{%
 .1 driver.
 .2 map.
 .3 route.
 .4 data\_modality.
 .5 camera.
 .6 weather.
 .7 frame.
}

\chA{In addition to the dataset itself, a separate directory named \texttt{miscellaneous} contains the scripts used to generate the driving routes, including the corresponding voice instructions in three languages. This material supports the reproducibility of the experiments and also enables researchers to go beyond the current dataset by extending the scenarios or introducing their own agents.}

\section{Examples of Application}\label{sc:Examples}

The LAIA dataset enables a wide range of applications aimed at advancing both the performance and interpretability of AI-based driving systems. In particular, the availability of synchronized gaze data, control signals, and rich scene annotations opens new directions for cognitive modeling and human-centered AI development. Next, we outline several representative use cases. \chA{From the perspective of gaze data alone, the dataset supports the training and evaluation of gaze prediction models. In addition, LAIA enables systematic comparison between the attention of AI drivers and that of human drivers, which can help assess whether models focus on semantically meaningful regions such as pedestrians, vehicles, or traffic signs, or making it possible to assess whether the models focus on the same scene elements that human drivers consider relevant, ultimately fostering the development of more explainable AI systems. The deliberate design of specific driving events along the routes also makes it possible to study how gaze patterns correlate with driving maneuvers in response to particular situations, such as a pedestrian emerging unexpectedly or a vehicle cutting in, thus supporting research on driving behavior modeling and prediction. Finally, by combining human attention with rich ground-truth annotations, the dataset provides a valuable benchmark for evaluating perception models under diverse weather conditions, viewpoints, and occlusion levels.}
These applications demonstrate the versatility of LAIA as a resource for 
improving autonomous driving systems.

\section{Conclusion}\label{sc:conclusion}

\chA{We presented LAIA, a novel dataset designed to support the development of attention-aware end-to-end autonomous driving systems. By combining raw sensory input, control actions, and high-resolution human gaze data in a wide variety of simulated scenarios, LAIA offers a unique resource for studying the interplay between perception, decision-making, and visual attention.

A key strength of LAIA lies in its deliberate design choices regarding data density and repeatability. Rather than maximizing the number of distinct environments, the dataset prioritizes depth over breadth: 44 participants drove through the same carefully crafted routes across a reduced set of towns, ensuring that each scenario was experienced by a large and diverse group of drivers. This design yields a high degree of within-scenario variability in human attention and driving behavior, since individual differences in gaze patterns, reaction times, and control responses can be directly compared under identical environmental conditions. The result is a statistically robust foundation for learning-based models, where the same scene is observed through the eyes of dozens of different drivers spanning a range of ages, experience levels, and driving styles. With over 15 hours of recorded driving synchronized with eye-tracking data, control signals, and rich ground-truth annotations, LAIA provides the repetition density that allows a robust comparison between AI and human drivers.

Furthermore, the six weather variations applied uniformly across all sequences add an orthogonal axis of robustness, ensuring that attention patterns are available not only across participants but also across lighting and meteorological conditions for the same underlying scene geometry. Combined with the controlled placement of driving events — designed specifically to elicit natural human reactions — this structure allows researchers to study not just where drivers look on average, but how attention shifts and varies in response to specific, reproducible stimuli.

The dataset enables new research directions focused on improving the transparency, interpretability, and cognitive alignment of AI drivers. By grounding end-to-end model development in the rich, repeated, and diverse human attention signals that LAIA provides, the autonomous driving community gains a tool designed not only for training, but for rigorous evaluation and meaningful comparison of attention-aware systems under controlled and repeatable conditions.}

\bibliographystyle{IEEEtran}
\bibliography{bibliography}

\end{document}